\documentclass[]{spie}  

\usepackage{amsmath,amsfonts,amssymb}
\usepackage{graphicx}
\usepackage{setspace}
\usepackage{tocloft}
\usepackage{booktabs}
\DeclareGraphicsExtensions{.pdf,.jpg,.png}
\title{Automated soft tissue lesion detection and segmentation in digital mammography using a u-net deep learning network}

\author[a]{Timothy de Moor}
\author[a]{Alejandro Rodriguez-Ruiz}
\author[a]{Albert Gubern M\'erida}
\author[a]{Ritse Mann}
\author[a,b]{Jonas Teuwen}

\affil[a]{Radboud University Medical Center, Department of Radiology and Nuclear Medicine, Nijmegen, the Netherlands}
\affil[b]{Optics Research Group, Imaging Physics Department, Delft University of Technology, the Netherlands}

\authorinfo{Send correspondence to Jonas Teuwen: jonas.teuwen@radboudumc.nl.}

\cftpagenumbersoff{figure}
\cftpagenumbersoff{table} 
\begin{document} 
\maketitle

\begin{abstract}
Computer-aided detection or decision support systems aim to improve breast cancer screening programs by helping radiologists to evaluate digital mammography (DM) exams. Commonly such methods proceed in two steps: selection of candidate regions for malignancy, and later classification as either malignant or not. In this study, we present a candidate detection method based on deep learning to automatically detect and additionally segment soft tissue lesions in DM. A database of DM exams (mostly bilateral and two views) was collected from our institutional archive. In total, 7196 DM exams (28294 DM images) acquired with systems from three different vendors (General Electric, Siemens, Hologic) were collected, of which 2883 contained malignant lesions verified with histopathology. Data was randomly split on an exam level into training (50\%), validation (10\%) and testing (40\%) of deep neural network with u-net architecture. The u-net classifies the image but also provides lesion segmentation. Free receiver operating characteristic (FROC) analysis was used to evaluate the model, on an image and on an exam level. On an image level, a maximum sensitivity of 0.94 at 7.93 false positives (FP) per image was achieved. Similarly, per exam a maximum sensitivity of 0.98 at 7.81 FP per image was achieved. In conclusion, the method could be used as a candidate selection model with high accuracy and with the additional information of lesion segmentation.
\end{abstract}



\begin{spacing}{1}   

\section{Introduction}
Population-based breast cancer screening programs with mammography have proven
to reduce mortality and the morbidity associated with advanced stages of the
disease. Nevertheless, their effectiveness is subject of 
discussion due to possible overdiagnosis, high false positive rate, or
insufficient detection rate for dense breasts. Still, one of 
the main pitfalls is possible interpretation errors by the radiologists who have to
evaluate a large amount of mammograms with a very low prevalence of malignant
cases.

Several studies have shown that a significant number of the diagnosed cancers
were already visible on previous screening mammograms which were marked as
negative \cite{Bae2014,Hoff2012}. Additionally, 
there is a significant variability between readers in terms of both sensitivity
and specificity \cite{Smith2005}, and therefore combining 
assessments by two or more readers \cite{Karssemeijer2004} 
improves screening performance.

Computer-aided detection or decision support systems could improve the breast
cancer screening programs by helping radiologists. Unfortunately, several
studies have shown that radiologists do not improve their screening performance
when using computer-aided detection systems, mainly because of a low specificity
of these traditional systems\cite{Lehman2015}. 

Recent developments in machine learning algorithms have greatly improved the
performance of computer vision and models in medical imaging by using deep
learning neural networks. It can therefore be expected that
a new generation of computer-aided detection or diagnosis systems for digital
mammography (DM) could be reliable used by radiologists to improve the
efficiency of breast cancer screening programs.

Generally, computer-aided detection or decision support systems proceed in
two steps. In the first step, the whole mammogram is processed and regions of
interest are selected, the so-called candidate detector. The primary goal of
this step is to greatly reduce the number of search locations while achieving a
sensitivity near $100\%$. In the second step, the goal is to remove
the false positives, while keeping the true positives. Up to now, most reported
candidate selectors are not based on deep learning technology. In this paper we
propose a deep learning network as candidate selector, which automatically
detects and additionally segments malignant lesions in DM images.

\section{Materials and Methods}
\subsection{Patient population}\label{sec:data}
This study was conducted with anonymized data retrospectively collected from our
institutional archive. It was approved by the regional ethics board after
summary review, with waiver of a full review and informed consent.

DM exams from women attending the national screening program at our collaborator
institution, or our institution for diagnostic purposes between 2000 and 2016
were included.

All cases with biopsy-proven malignant soft tissue lesions were collected, while normal exams
were selected if they had at least two years of negative follow-up. This yielded
a total of 7196 DM exams, from which 2883 exams (42\%) contained a total of 3023
biopsy-verified malignant lesions. Most exams were bilateral and included two views
(cranio-caudal -CC- and medio-lateral oblique -MLO-), resulting in a total of
28924 images. The exact distribution is summarized in Table~\ref{table:table}.
\begin{table}[h!]
\centering
\caption{Distribution of the DM exams included in this study.}
\label{table:table}
\begin{tabular}{|l|l|l|l|}
  \hline
 {} & General Electric & Siemens & Hologic \\ \hline
 number of studies & 2248 & 1518 & 3430\\
 \,- normal images& 7771 & 5842 & 12288 \\
 \,- images with malignant lesion(s)& 1292 & 255 & 1476  \\ \hline
\end{tabular}
\end{table}

\subsection{Image acquisition and preprocessing}\label{sec:acquisition}
The images were acquired by four DM machines from three different vendors (Senographe 2000D and Senographe DS, General Electric, USA; Mammomat
Inspiration, Siemens, Germany; Selenia Dimensions, Hologic, USA;). Distributions
are shown in Table~\ref{table:table}.

\begin{figure}[h!]
  \includegraphics[width=1\linewidth]{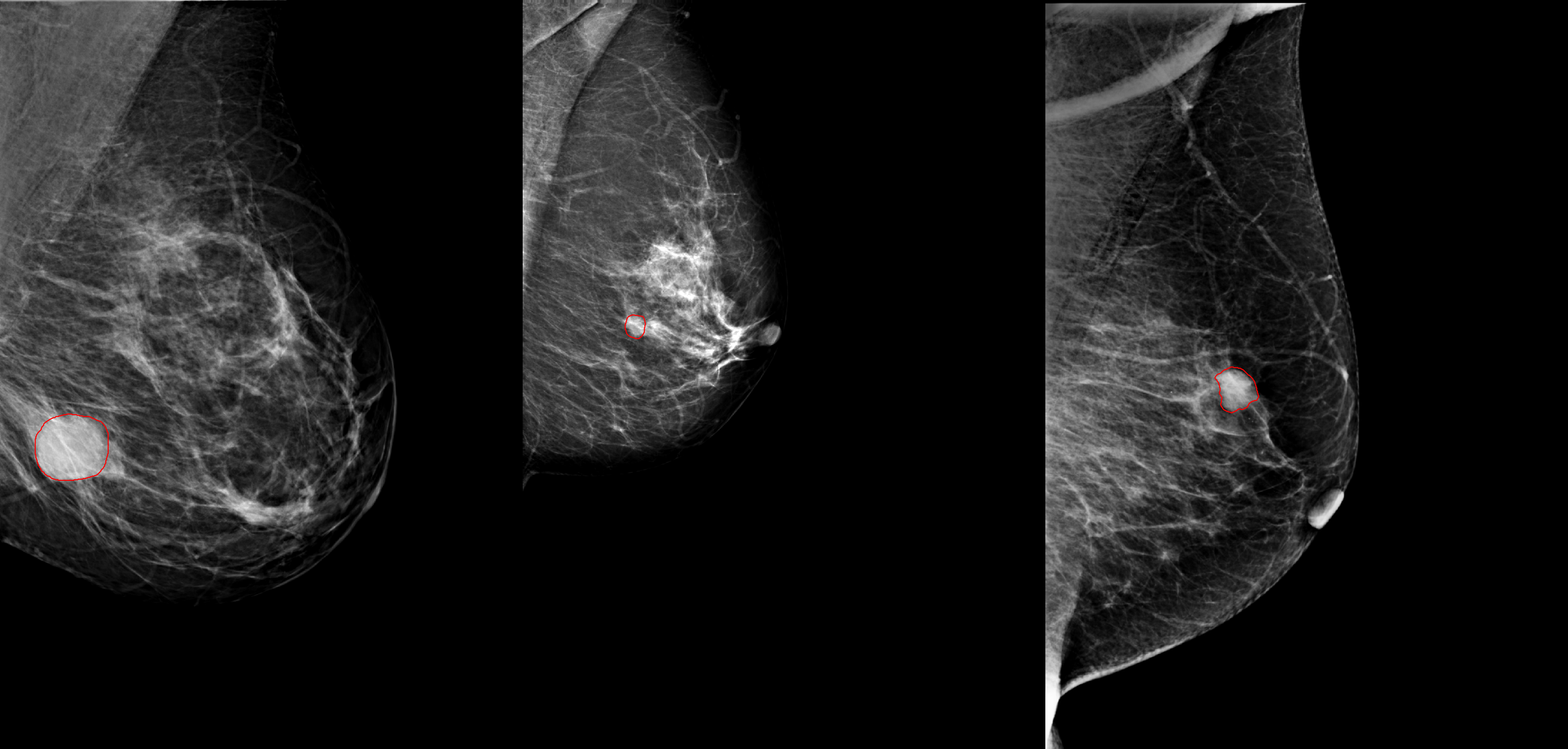}
  \caption{Examples of DM images acquired with machines from three vendors:
    General Electric (left), Siemens (middle) and Hologic (right); after
    preprocessing. Biopsy-proven lesions are outlined.}
  \label{fig:example-normalized}
\end{figure}

Images were preprocessed in three steps. First, an energy band normalization
technique was applied \cite{Philipsen2015}. To homogenize the pixel size
across different vendors, images were downscaled to $200 \mu\text{m}$ after
applying a Gaussian filter. Finally, pixel values were scaled to the range $[0,
1]$. Examples of preprocessed images with outlined lesions are shown in
Figure~\ref{fig:example-normalized}.

\subsection{Ground truth labeling and segmentation}\label{sec:annotations}
All cases with malignant lesions were verified by histopathology and manually annotated and segmented under
the supervision of an expert breast radiologist, with access to other breast
imaging exams, radiological and histopathological reports. More than $99\%$ of
the lesions had a maximal bounding box diameter of $5.7 \text{cm}$ and on average
had a height/width ratio of about $1$. The negative cases were verified by at
least two years of negative follow-up imaging exams.

\subsection{Deep learning network and training}\label{sec:unet}
\begin{figure}[h!]
  \includegraphics[width=0.8\linewidth]{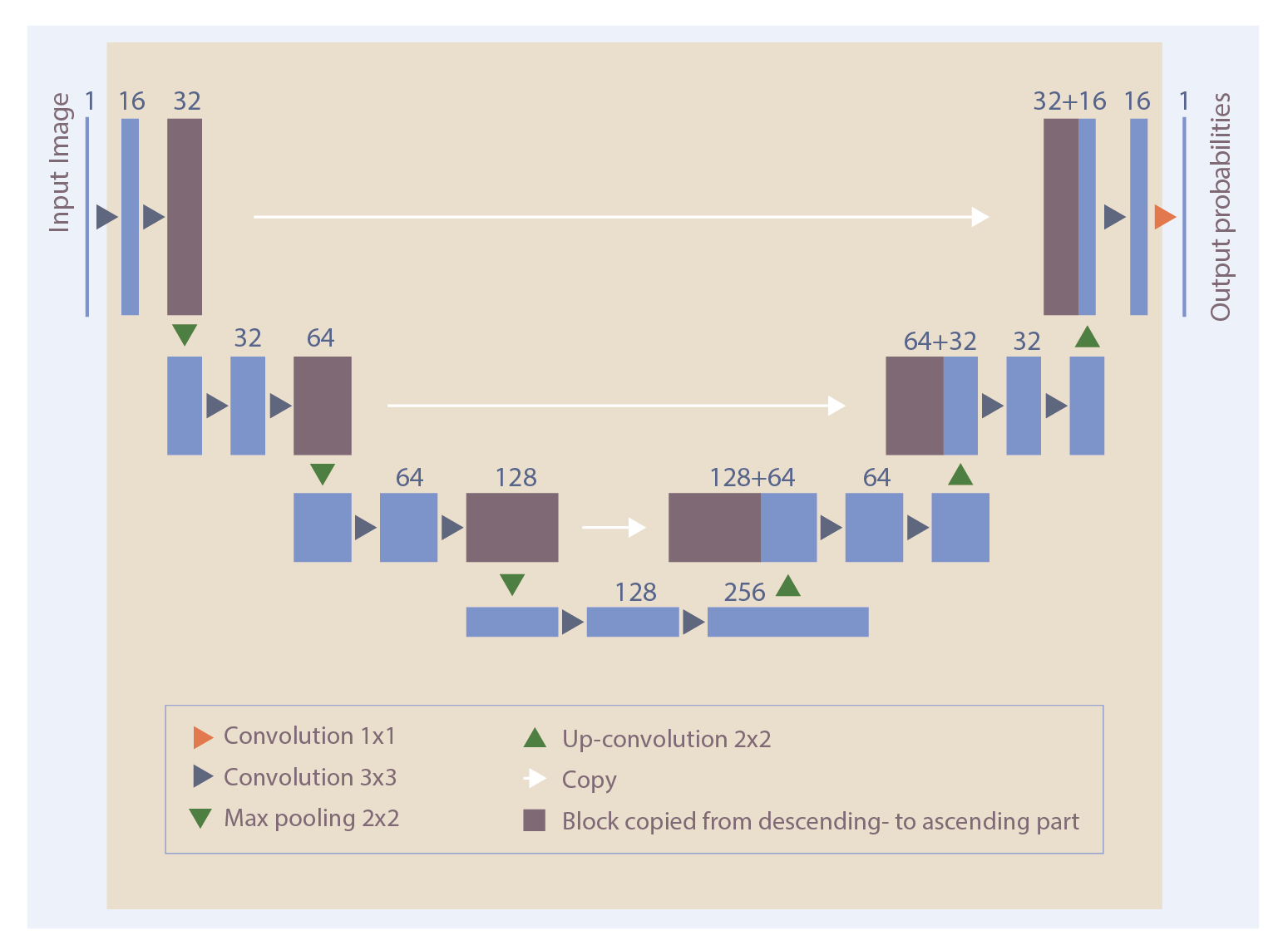}
  \caption{The u-net architecture, where we have doubled the number of filters
    in the each block and applied batch normalization. Each
    convolution $3 \times 3$ was followed by a rectified linear unit activation
    layer, while the final convolution $1 \times 1$ was followed by a pixel-wise
    sigmoid activation layer.}
  \label{fig:unet}
\end{figure}

The goal was to detect and segment the malignant lesions. For that, a deep learning convolutional neural
network with modified u-net architecture \cite{Ronneberger2015} was 
employed (detailed in Figure~\ref{fig:unet}) and trained on data from all three vendors.

The u-net worked on a patch level ($344 \times 344$ pixels = $6.88 \times 6.88
\text{cm}$, considered to provide enough context to discriminate soft tissue
lesions). A u-net is a so-called fully convolutional network, so it is agnostic
to the size of the input, so we can evaluate the trained network on complete
mammograms. Positive samples were extracted around the center-of-mass of the annotated lesion and
zero-padded if necessary. Negative patches were randomly selected patches within
the breast from the normal cases. During each epoch all positive samples were
presented and an equal number of randomly selected negatives. Data was split on
a DM exam level to avoid bias: $50\%$ of the data was used for training of the
u-net model, $10\%$ for the validation, and $40\%$ for independent test. 

The model was trained using stochastic gradient descent with momentum (0.9)
using a weighted logistic loss function with weight $0.25$ for the negative
samples. The learning rate, initially $0.005$ was reduced by a factor of $2$ if
the validation loss did not improve for at least 5 epochs. Up-down and
left-right flips were used as data augmentations.

\subsection{Performance analysis: FROC}\label{sec:unet}
The performance of the model was evaluated with free receiver operating
characteristic (FROC) analysis, on a region and case basis.

Given a mammogram, the deep neural network produces a probability map where each
pixel represents the probability that this pixel belongs to a suspicious lesion.

Before we proceed with the FROC analysis where we compare predicted locations
with the ground truth, we need to convert this probability map to a list
of candidate locations and the ground truth segmentation to such a list as well.

To convert the probability map to a list of candidates, we first binarize the
map by thresholding at $0.5$ which is the value which gave good performance on
the validation and training sets. On this map we compute all connected
components. Given these components, we generate candidates at threshold $T$ by
finding all points which have probability values above $T$. To reduce the number
of candidates, we cluster all candidates within a radius of $1.5\text{cm}$. The
ground truth segmentation are converted into coordinates by computing the center
of mass for each lesion.

Given the candidates and the ground truth, we can compute two types of FROC
curves: the image-based and the exam-based FROC curve. In the FROC analysis,
lesion is said to be correctly predicted if there is a candidate within
$1.5\text{cm}$ of the center-of-mass of the lesion.

In both FROCs we plot the average number of false positives per image (FP/image)
against the true-positive rate (TPR). A point on the FROC is computed as
follows: for each threshold $T$ we count the number of false positives on the
normals as we can guarantee there are no lesions there. In the image-based case
we compute the average TPR per image, where the TPR per image is the ratio of
the number of lesions correctly predicted in this image. The exam-based approach
gives a TPR of $1$ if at least one of the lesions is correctly predicted in the
study, and else $0$.

\section{Results}
\begin{figure}[h!]
\includegraphics[width=1\linewidth]{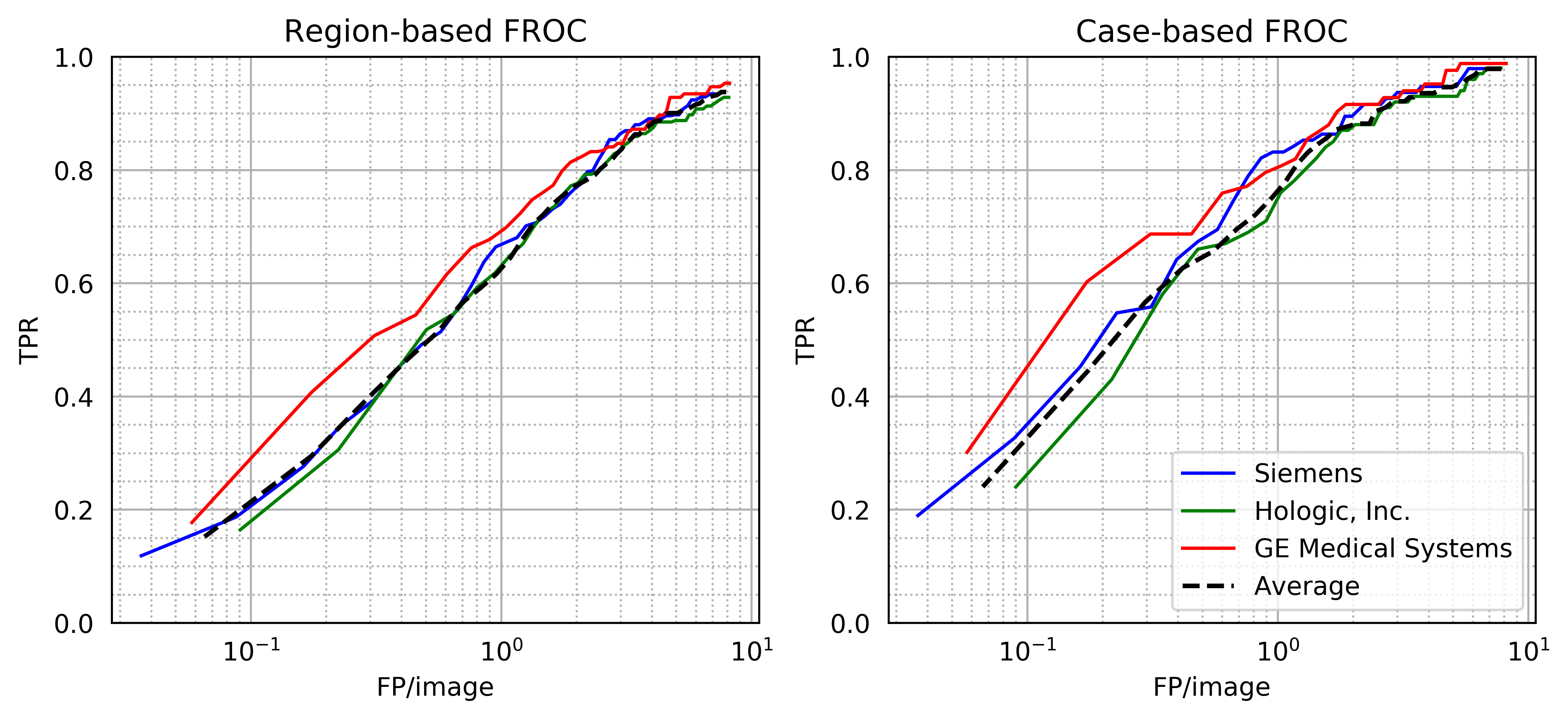}
\caption{Image-based and exam-based FROC performance of the u-net. We reach a maximum
  sensitivity of $0.94$ and $0.98$ at a false positive rate per image of $7.93$
  and $7.81$ for the image-based and exam-based respectively.}
\label{fig:FROC}
\end{figure}

\begin{figure}[h!]
\includegraphics[width=0.95\linewidth]{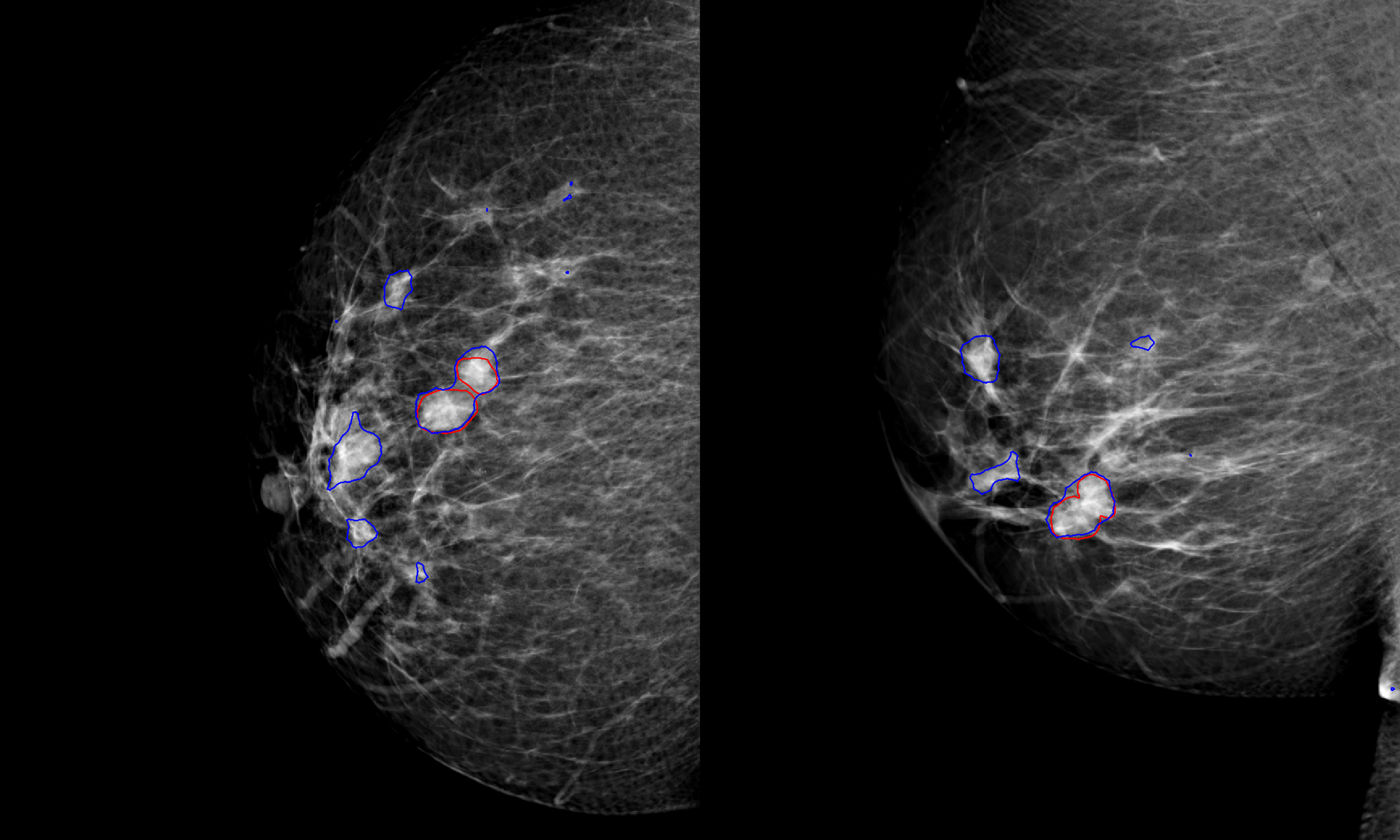}
\caption{Example of a case with biopsy-proven invasive lobular carcinoma (ILC),
  where a few false positive candidates are predicted. The predictions are
  delineated in blue, whereas the ground truth is delineated in red. We can see
  that the annotation is split in two parts, while the prediction connects these
  regions. Given the intra-reader variability of the segmentations, this is to
  be expected and is not an issue for our analysis.}
\label{fig:example1}
\end{figure}

\begin{figure}[h!]
\includegraphics[width=0.95\linewidth]{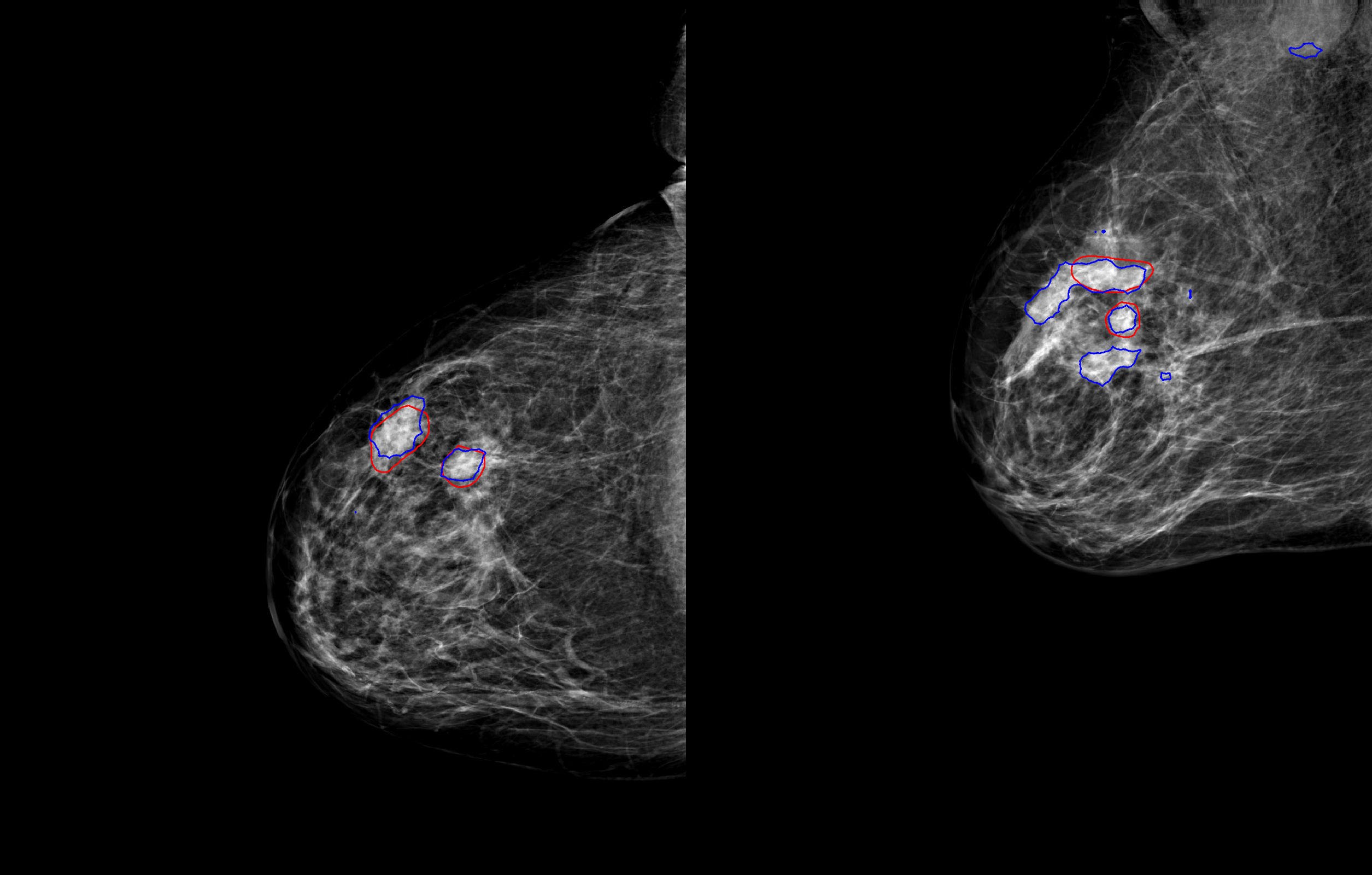}
\caption{Example of a case with biopsy-proven invasive lobular carcinoma (ILC),
  which is properly segmented. The predictions are delineated in blue, whereas
  the ground truth is delineated in red.}
\label{fig:example2}
\end{figure}

\begin{figure}
\includegraphics[width=0.95\linewidth]{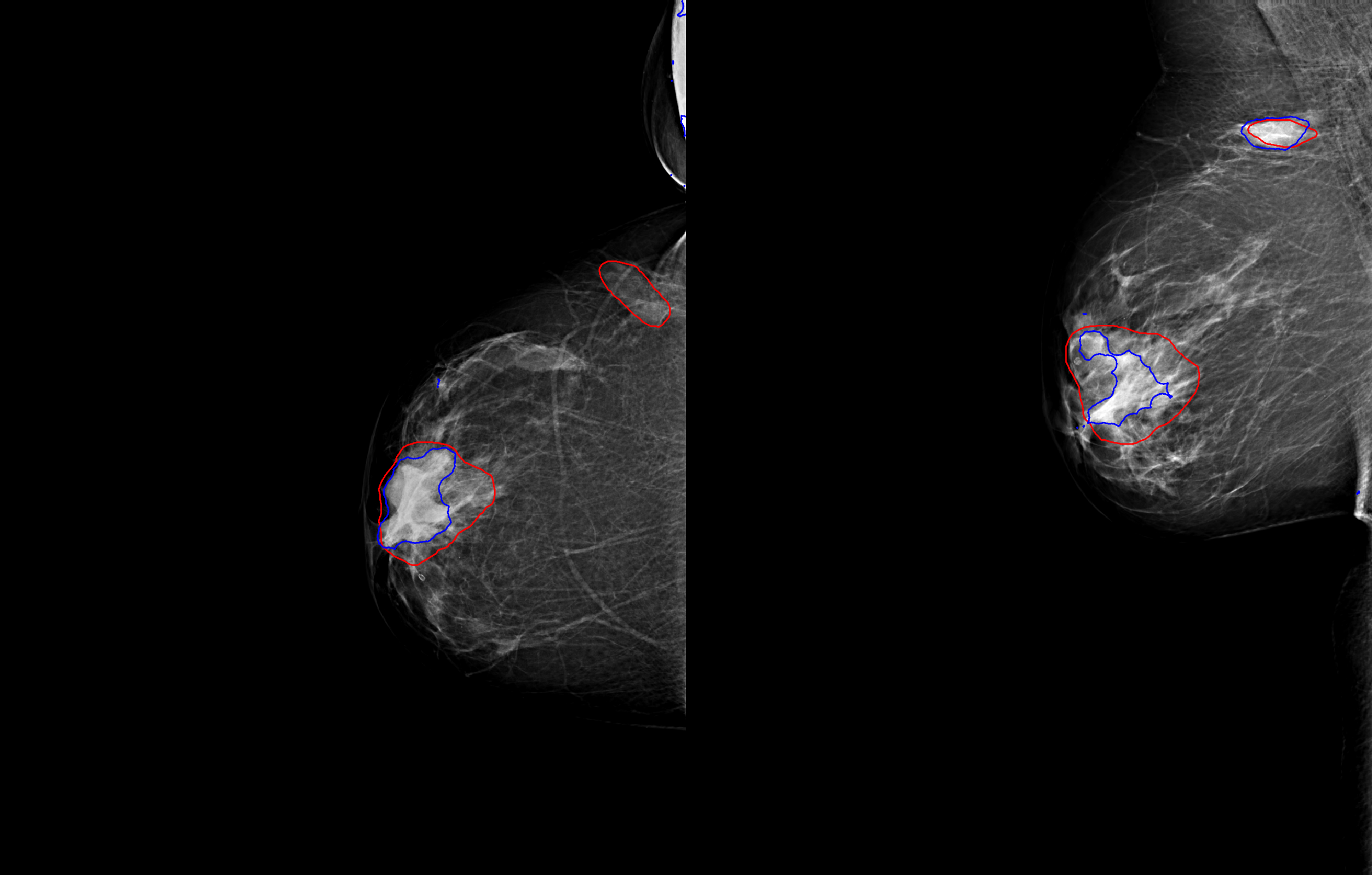}
\caption{Example of a case with non-Hodgkin b-cell lymphoma, where the lesions
  are properly detected in the MLO view, but not in the CC view. The predictions
  are delineated in blue, whereas the ground truth is delineated in red. The
  network did not have the information that both views belong together, while
  the annotator did.}
\label{fig:example3}
\end{figure}

The resulting FROC curves are given in Figure~\ref{fig:FROC}. For the image-based
FROC we obtain a maximum sensitivity of $0.94$ with a false positive rate per
image of $7.93$ at a threshold of $0.5$ (which is the lowest threshold where the
sensitivity did not further improve). In the exam-based FROC, the
maximum sensitivity is $0.98$ with a false positive rate per image of $7.81$.
Several examples are shown in Figures \ref{fig:example1}-\ref{fig:example3}.


\section{Conclusions}
In this abstract we presented a method which detects and segments soft tissue lesions in
digital mammography, and could be used as a candidate detection model for
automated detection and diagnosis computer systems. Therefore, it is important
that such candidate model can detect all exams with cancer lesions. Our model
achieved an acceptable performance of $0.98$ sensitivity with a false positive
rate per image of $7.81$ in the exam-based FROC.

Compared to the classical candidate detection techniques, we also provide a
segmentation of the lesion. This allows to study the temporal changes, such as
growth or morphological modifications over time. Furthermore, if we have information
available of the shape of the projection of the lesion, this might improve the
correlation of candidates between the MLO and CC views more precisely.

One of the main limitations of our study is that we only studied soft tissue
lesions. A model that can also detect and segment calcifications would be of
great added value and the feasibility of using a u-net deep learning
architecture for this task should be explored. Also, including data from more
manufacturers or some other normalization techniques could allow to reduce the
performance variability across vendors. Finally, introducing benign lesions to provide the model with more information should also be explored, in order to reduce the number of
false positive assessments.

Another topic of future work is to study whether this two-dimensional detection
and segmentation model can be applied in digital breast tomosynthesis (slice by
slice basis) or synthetic mammography images, both of which are often combined
with DM as a breast cancer screening protocol. Having a robust candidate
detection for all of these three types of images would be beneficial to also
have a robust computer system that can correlate information across modalities. 

\newpage
\bibliography{bibliography}   
\bibliographystyle{spiejour}   

\end{spacing}
\end{document}